# An N-dimensional approach towards object based classification of remotely sensed imagery

Arun P.V[1]        Dr. S.K. Katiyar

Dept. Of Civil
MANIT-Bhopal, India
Ph: +914828249999

## Abstract

Remote sensing techniques are widely used for land cover classification and urban analysis. The availability of high resolution remote sensing imagery limits the level of classification accuracy attainable from pixel-based approach. In this paper object-based classification scheme based on a hierarchical support vector machine is introduced. By combining spatial and spectral information, the amount of overlap between classes can be decreased; thereby yielding higher classification accuracy and more accurate land cover maps. We have adopted certain automatic approaches based on the advanced techniques as Cellular automata and Genetic Algorithm for kernel and tuning parameter selection. Performance evaluation of the proposed methodology in comparison with the existing approaches is performed with reference to the Bhopal city study area.

## I.     Introduction

Land cover plays a pivotal role in impacting and linking many parts of the human and physical environments, hence monitoring of land cover and its changes has great significance.

---

[1] Email:arunpv2601@gmail.com

Remote sensing techniques are gaining more and more importance for land cover classification and urban analysis. The accuracy of pixel based classification approaches are affected by the increase in resolution of images and object based approaches are devised for improving the performance (Vapnik et al., 1998). The availability of high resolution satellite images have popularised the object based classification and literature suggests a great deal of advanced methodologies for the purpose (Nghi et al., 2008). The spectral and spatial information can be combined to increase the seperability between classes to yield higher classification accuracy (Gregoire et al., 2004).

Support Vector Machines (SVM) technique (Hosseini et al., 2009) is a relatively recent generation of classifiers based on advances in statistical learning theory (Burges et al., 1998). The SVM methodologies are particularly appropriate for remote sensing data analysis and have been applied to the classification of multispectral (Yanfeng et al., 2008) and hyper spectral (Lu et al.,2011; Melgani et al.,2008) images. The technique constitute of finding the optimal separation between the classes in an n-dimensional plane. This technique uses kernel method to project linearly inseparable data to a higher dimension space using appropriate kernels. Kernel methods have useful properties when dealing with low number of (potentially high dimensional) training samples, the presence of heterogeneous multimodalities, and different noise sources in the data (Chi-Hoon et al., 2005). The kernel method may perform class separation even with means very close to each other with a small number of training samples. Every function that satisfies Mercer's conditions (Hosseini et al.,2009) may be considered as an eligible kernel. The existing SVM approaches adopts separability measures based on dot product or geometric distance between vectors without taking the spectral meaning and behaviour in to consideration (Lennon et al.,2007).

As SVMs can adequately classify any data in a higher dimensional feature space with a limited number of training datasets, it overcomes the Hughes Phenomenon (Lennon et al., 2007). In fact, kernel methods have improved results of parametric linear methods and neural networks in applications such as natural resource control, detection and monitoring of anthropic infrastructures, agriculture inventorying, feature extraction etc (Nghi et al., 2008). Even if an object is observed with several illumination conditions, its spectral signature remains the same and has to be classified in the same way. Mercier et.al(2004 ) proposed the linear mixing of quadratic with spectral kernels (Spectral Information Divergence& Spectral angle based) to achieve better classification results as compared to the statistical based approaches.

The SVM is an independent and identically distributed classifier that does not consider interactions in the labels of adjacent data points but have the appealing generalization properties (Lee et al., 2005). The advanced classifiers as Markov Random Field (MRF) and Conditional Random Field (CRF) are proposed to augment the performance of SVM (known as SVRF) by taking into account spatial class dependencies (Lee et al.,2006). Conditional Random Fields which are an extension of the Markov Random Fields, can better model spatial dependencies between labels and features by taking in to consideration of the adjacency interactions. The Support Vector Random Field (SVRF) that combines CRF and SVM is found to outperform SVMs and DRFs (Farid et al., 2004). The Support Vector Random Field model is robust to class imbalance, can be efficiently trained, converges quickly during inference, and can trivially be augmented with kernel functions to improve results (Lee et al., 2005). The SVRF can attain the appealing generalization properties of SVMs and the ability to model different types of spatial dependencies of CRFs.

Schnitzspan et.al (2008) proposed a hierarchical support vector random field based approach that combines the power of global feature-based approaches with the flexibility of local feature-based methods. Authors have incorporated SVMs and multiple layers of CRFs in one consistent framework in order to automatically learn the trade off and the optimal interplay between local, semi-local and global feature contributions. Gustavo et.al (2006) suggested soft classification of hyper spectral imagery by incorporating the spatial and spectral information using the composite kernel based SVM.

In this paper we adopt a hierarchical SVRF model for producing multiclass SVMs for object based classification and compare various kernel methods suitable for remote sensing with reference to the available sensor data. We have adopted certain automatic approaches based on the advanced techniques as Cellular automata and Genetic Algorithm for kernel and tuning parameter selection.

## 2. Mathematical formulation & methodology

*2.1 SVM*

The SVM based classifier is a separating hyper plane that is defined by the most important training points (support vectors). Given a set of training pixels $\mathbf{x}_i \in \mathbf{R}^d$ and output classes $y_i \in \{-1, 1\}$, SVM utilizes a hyper plane to linearly separate between the two classes. The hyper plane can be specified as an optimization problem as

$$\Phi(\mathbf{w}) = \|\mathbf{w}\|^2 \text{ s.t. for all } (\mathbf{x}_i, y_i), i=1..n : \quad y_i(\mathbf{w}^T \mathbf{x}_i + b) \geq 1$$

Quadratic optimization algorithms can identify the support vectors with non-zero Lagrangian multipliers $α_i$. The quadratic optimization can be obtained for the roots as $α_1…α_N$ such that

*$Q(α) = Σα_i - ½ΣΣα_iα_jy_iy_jx_i^Tx_j$ is maximized and  $Σα_iy_i = 0$ , $0 ≤ α_i ≤ C$ for all $α_i$ .*

By methods like quadric optimization, unknown can be obtained, and given an input pixel P, its SVM output can be written by, *$g(P) = Σ_i α_i\, y_i(P, P_i) + b$*

The Kernel functions are used for projecting the inseparable data values to higher dimension and hence the output can be denoted as $g(P) = Σ_i α_i\, y_i K(P, P_i) + b$ , where $K(P, P_i)$ is the Kernel function for a given input pixel P and support vector $P_i$. The posterior probability of each pixel is iteratively calculated for multiclass as discussed in (Yanfeng et al.,2008).

*2.2 kernels*

The composite kernel concept is used to incorporate spectral and spatial information, given $X = \{x_1, x_2, ..x_m\}^T$ be the spectral characteristics of an M-band multispectral imagery and $Y = \{y_1, y_2, ..y_n\}^M$ be the spatial characteristics, then the possible spectral and spatial kernels can be denoted as $K_x(P, P_i) = \langle Φ(P), Φ(P_i) \rangle$ , $K_y(P, P_i) = \langle Ψ(P), Ψ(P_i) \rangle$ respectively. Preferably a weighted combination of the kernels are adopted as discussed in (Yanfeng et al.,2008) such that $K(P, P_i) = μK_x(P, P_i) + (1-μ)K_y(P, P_i)$ and the value of tuning parameter is adjusted accordingly.

*2.3 SVRF*

SVRF (Chi-Hoon et al., 2005)(Lee et al.,2005)(Lee et al.,2006) is a Discrete Random Field (DRF) based extension for SVM, constituting of observation-matching potential function and the local-consistency potential function. The observation-matching function captures relationships between the observations and the class labels, while the local-consistency function models relationships between the labels of neighbouring data points and the observations at data points.

$$P(Y \mid X) = \frac{1}{Z} \exp\left\{ \sum_{i \in S} \log(O(y_i, \Gamma_i(X))) + \sum_{i \in S} \sum_{j \in N_i} V(y_i, y_j, X) \right\}$$

In this formulation, $\Gamma_i(X)$ is a function that computes features from the observations $X$ for location $i$, $O(y_i, i(X))$ is an SVM-based Observation-Matching potential and $V(y_i, y_j, X)$ is a (modified) DRF pair wise potential.

### 2.4 Proposed algorithm

The SVRF is trained to generate object based CA rules which are used to incorporate contextual information to the kernels and are also used for tuning parameter selection. The tuning parameters as well as kernels are selected using Genetic Algorithm and Cellular Automata Techniques. The input data is initially segmented to determine the compatibility of objects with reference to trained data and further posterior probabilities are calculated. The schematic representation of the algorithm is as given in the (Figure 1).

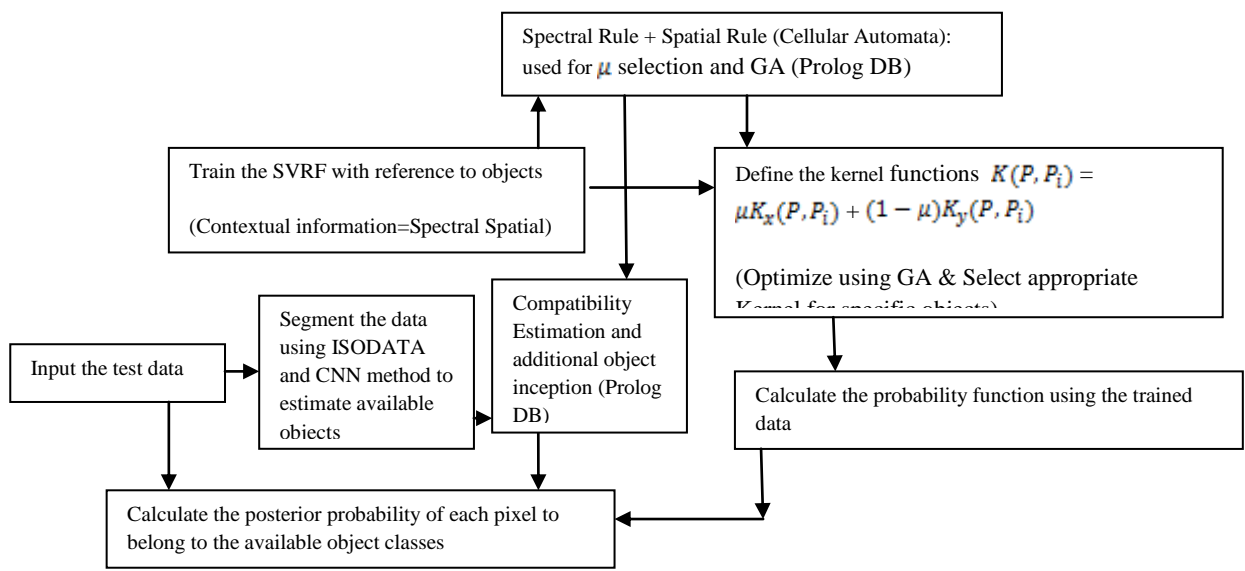

Figure 1. Proposed algorithm

## 3. Experiments

*3.1 Data*

SVRF classifications have been applied to the multispectral image from the LISS III and LISS IV sensor of Indian Remote Sensing Satellites and details are as given in (Table 1). The image has been geo referenced using ERDAS 9.1 and has been sub set for the Bhopal Area.

Table 1. Details of experimental data

| S.No. | Imaging sensor | Spatial resolution(m) | Satellite | Area | Date of Acquisition |
|---|---|---|---|---|---|
| 1 | LISS-III | 23.5 | IRS-P6 | Bhopal(India) | 5$^{th}$ April 2009 |
| 2 | LISS-IV | 5.6 | IRS-P6 | Bhopal(India) | 16$^{th}$ March 2010 |

*3.2 Implementation*

The algorithms are implemented in MatLab and various kernels are analysed and spectral information is encoded using cellular automata technique. The results of implementations are evaluated using cross validation technique (Melgani et al.,2008) and the ground truth test data. Certain accuracy criteria such as *Overall Accuracy* and *Kappa* Coefficient of agreement (Tan et al., 2011) are estimated using confusion matrix and the accuracy analysis is done using Matlab and ERDAS. The procedure of accuracy estimation is as summarised in (Figure 2).

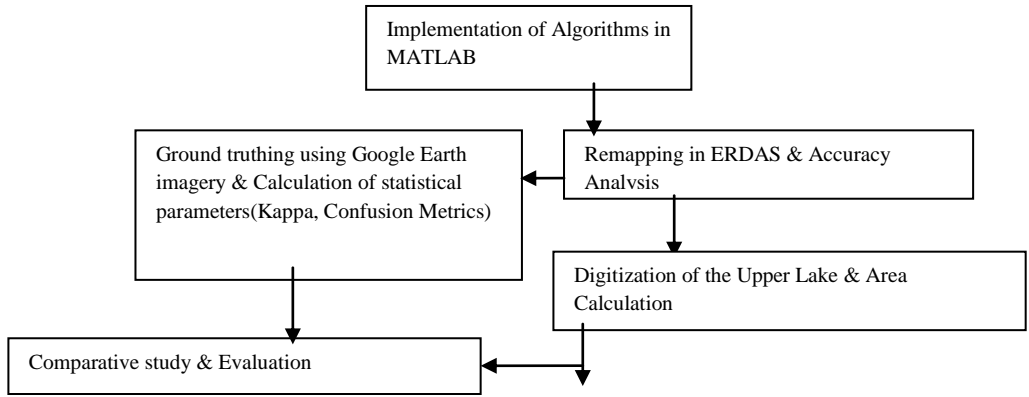

Figure 2. Accuracy Analysis

*3.3 Results and discussions*

The investigations of this research work revealed that the use of spectral knowledge into SVRF classification reduces false alarms for thematic classification. For instance, recreational forest area (Van Vihar national park- Bhopal), which is difficult to classify since trees are small and there is a lot of shadows, has been correctly classified with SVRF approach. The efficiency of the traditional classifying approaches with reference to the SVRF approach has been evaluated using the various statistical measures and the results are as summarised in (Table 2). The ground truthing is done with reference to the Google earth and Differential Global Positioning System (DGPS) survey over the study area using Trimble R3 DGPS equipment.

Table 2. Results of Accuracy Analysis

| S.No | Sensor | Methodology | Kappa statistics | Overall Accuracy (%) |
|---|---|---|---|---|
| 1 | LISS 3 | Mahalanobis | 0.93 | 93.13 |
| 2 | LISS 3 | Minimum Distance | 0.92 | 94.58 |
| 3 | LISS 3 | Maximum Likelihood | 0.96 | 96.83 |
| 4 | LISS 3 | Parrellelepipid | 0.95 | 96.81 |
| 5 | LISS 3 | Feature Space | 0.97 | 95.15 |

| S.No | Sensor | Methodology | | |
|---|---|---|---|---|
| 6 | LISS 3 | SVM(Spectral & spatial factor considered) | 0.99 | 97.13 |
| 7 | LISS 3 | SVRF(Spectral & spatial factor considered) | 0.99 | 97.51 |
| 8 | LISS 4 | Mahalanobis | 0.90 | 91.40 |
| 9 | LISS 4 | Minimum Distance | 0.91 | 93.00 |
| 10 | LISS 4 | Maximum Likelihood | 0.94 | 94.80 |
| 11 | LISS 4 | Parrellelepipid | 0.93 | 94.62 |
| 12 | LISS 4 | Feature Space | 0.94 | 95.3 |
| 13 | LISS 4 | SVM(Spectral & spatial factor considered) | 0.98 | 96.84 |
| 14 | LISS 4 | SVRF(Spectral & spatial factor considered) | 0.99 | 97.2 |

The investigation results reveal that the classification accuracy of the traditional methods is affected by the increase in the resolution of satellite images. Accuracy of the SVRF based methodologies is found to be comparatively stable over the change in resolution. The performances of these methodologies are also evaluated by comparing the areal extents of various features. The features having well defined geometry like lakes, parks etc are selected for the comparative analysis. The original surface areas of the features are calculated by manual digitization using ERDAS and comparative the results are presented in the (Table 3). Comparative analyses of the areal extents also indicate that the SVRF approach yields better results compared to the other methods. The Van Vihar national park which is a recreational forest area can be distinguished by using the SVM based approaches and this indicates the superiority of SVM approaches for object based classification.

Table 3. Comparison of the geographical extent of various features

| S.No | Sensor | Feature | Reference Area(km²) | Methodology | Areal Extent(km²) |
|---|---|---|---|---|---|
| | | | | Mahalanobis | 25.42 |
| | | | | Minimum Distance | 24.31 |
| | | | | Maximum Likelihood | 27.37 |
| 1 | LISS3 | Lake | 32.5 | Parallelepiped | 28.58 |
| | | | | Feature Space | 26.82 |
| | | | | SVM(Spectral & spatial factor considered) | 28.71 |
| | | | | SVRF(Spectral & spatial factor considered) | 30.72 |
| 2 | LISS3 | Parks | 2.13 | Mahalanobis | 0.82 |
| | | | | Minimum Distance | 0.89 |
| | | | | Maximum Likelihood | 1.45 |
| | | | | Parallelepiped | 1.37 |
| | | | | Feature Space | 1.56 |
| | | | | SVM(Spectral & spatial factor considered) | 1.51 |
| | | | | SVRF(Spectral & spatial factor considered) | 1.65 |
| 3 | LISS3 | Artificial Forest area (Vanvihar) | 4.41 | Mahalanobis | -- |
| | | | | Minimum Distance | -- |
| | | | | Maximum Likelihood | -- |
| | | | | Parallelepiped | -- |
| | | | | Feature Space | -- |
| | | | | SVM(Spectral & spatial factor considered) | 3.52 |
| | | | | SVRF(Spectral & spatial factor considered) | 2.61 |
| 4 | LISS4 | Lake | 32.81 | Mahalanobis | 24.31 |
| | | | | Minimum Distance | 23.40 |
| | | | | Maximum Likelihood | 25.12 |
| | | | | Parallelepiped | 26.24 |
| | | | | Feature Space | 27.17 |
| | | | | SVM(Spectral & spatial factor considered) | 28.63 |
| | | | | SVRF(Spectral & spatial factor considered) | 30.08 |
| 5 | LISS3 | Parks | 2.37 | Mahalanobis | 0.51 |
| | | | | Minimum Distance | 0.72 |
| | | | | Maximum Likelihood | 1.53 |
| | | | | Parallelepiped | 1.14 |
| | | | | Feature Space | 1.46 |
| | | | | SVM(Spectral & spatial factor considered) | 1.63 |
| | | | | SVRF(Spectral & spatial factor considered) | 1.71 |
| 6 | LISS3 | Artificial Forest area (Vanvihar) | 3.95 | Mahalanobis | -- |
| | | | | Minimum Distance | -- |
| | | | | Maximum Likelihood | -- |
| | | | | Parallelepiped | 1.81 |
| | | | | Feature Space | -- |
| | | | | SVM(Spectral & spatial factor considered) | 3.42 |
| | | | | SVRF(Spectral & spatial factor considered) | 3.62 |

The classified results for the LISS 3 imagery using various methodologies are as given in (Figure 3) and visual interpretation also reveals the accuracy of SVRF based methodology.

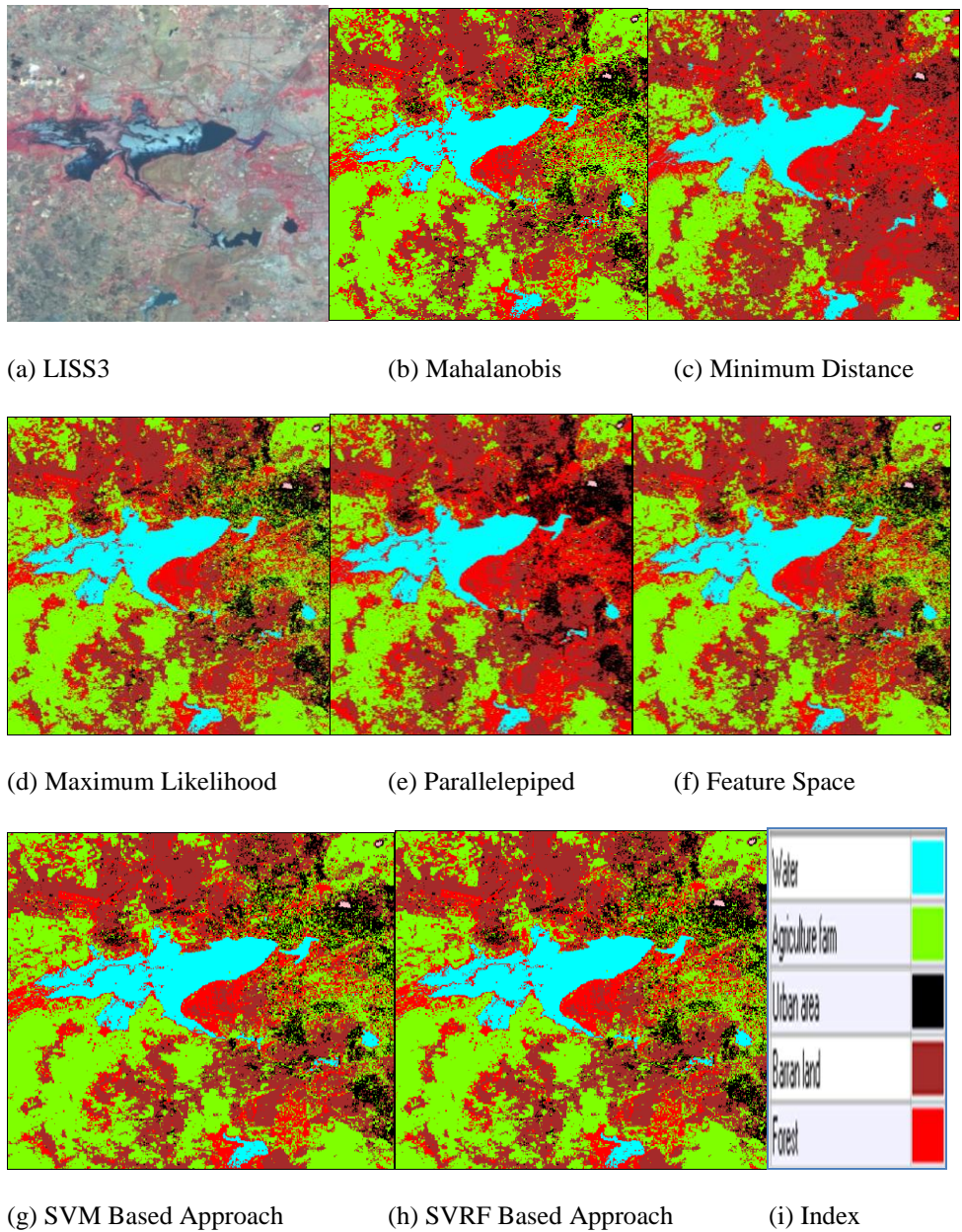

(a) LISS3            (b) Mahalanobis            (c) Minimum Distance

(d) Maximum Likelihood      (e) Parallelepiped            (f) Feature Space

(g) SVM Based Approach      (h) SVRF Based Approach      (i) Index

Figure 3. Visual comparison of different classification methods for LISS3 sensor imagery

## 4. Conclusion

SVM is found to give better results when augmented by the probabilistic approaches like CRF which considers the spatial dependencies of the classes. The investigation revealed that use of spectral knowledge into SVRF classification reduces false alarms for thematic classification. The proposed use of CA for the incorporation of rules and GA for the optimized selection found to yield better results. SVRF based approach is found to outperform the contemporary methods and can be made semi supervised by enhancing with Learning Automata.

## 5. References

Burges C. J., Fayyad U., "A Tutorial on Support Vector Machines for Pattern Recognition," *Data mining and knowledge discovery*, Ed. Kluwer Academic, pp.1-43, 1998.

Chi-Hoon Lee, Russell Greiner, Mark Schmidt, "Support Vector Random Fields for Spatial Classification," *PKDD 2005*, pp.121-132,2005.

Farid Melgani., "Classification of Hyperspectral Remote Sensing Images With Support Vector Machines," *IEEE Transactions On Geosciences and Remote Sensing,* Vol. 42, no. 8, pp.234, 2004.

Gregoire Mercier and Marc Lennon, "Support Vector Machines for Hyperspectral Image Classification with Spectral-based kernels,*" IEEE-Transactions of Geo Science and Remote Sensing, Vol.45, no.3, pp*. 123-130, 2003.

Gustavo C.V., Luis el G.C., "Composite Kernels for Hyperspectral Image Classification," *IEEE Geo science Remote Sensing Letters*, vol.3, no.1, pp. 93-97, 2006.

Hosseini R.S., Homayouni S., "A SVMS-based hyperspectral data classification algorithm in a similarity space," *Work Shop on Hyperspectral Image and Signal Processing: Evolution in Remote Sensing, 2009: WHISPERS '09*, vol.1, pp.1-4.

Huang C., Davis L. S., and Townshend J. R. G., "An assessment of support vector machines for land cover classification," *Int. J. Remote sensing*, vol. 23, no. 4, pp. 725–749, 2002.

Lee C. H., Schmidt M., Murtha A., Bistritz A., Sander J., and Greiner R., 2006, Segmenting brain tumor with conditional random fields and support vector machines, *Proc. Workshop Comp. Vis. Biomed. Image Appl.: Current Techniques Future Trends*, pp.469.

Lee, C., Schmidt M., Greiner R., "Support vector random fields for spatial classification," *9th European Conference on Principles and Practice of Knowledge Discovery in Databases (PKDD)*, Portugal, pp. 196, 2005.

Lennon M., Mercier G., and Hubert-Moy L., Classification of hyperspectral images with nonlinear filtering and support vector machines, *IGARSS*, Vol.22, no.2, pp. 141-147, 2007.

Lu D., Weng Q., "A survey of image classification methods and techniques for improving classification performance," *International Journal of Remote Sensing*, Vol.28 no.5, pp.823-870, 2011.

Melgani F., and Bruzzone L., "Support vector machines for classification of hyperspectral remote sensing images," *IGARSS*, Vol.23, pp. 234-240, 2008.

Nghi Dang Huu, Mai Luong Chi., "An object-oriented classification techniques for high resolution satellite imagery," *GeoInformatics for Spatial-Infrastructure Development in Earth and Allied Sciences (GIS-IDEAS)*, pp. 230-240, 2008.

Paul Schnitzspan, Mario Fritz, and Bernt Schiele., "Hierarchical Support Vector Random Fields: Joint Training to Combine Local and Global Features," *COMPUTER VISION – ECCV 2008 Lecture Notes in Computer Science*, Vol.5303/2008, pp. 527-540, 2008.

Tan K. C., Lim H. S., and Mat Jafri M. Z., "Comparison of Neural Network and Maximum Likelihood Classifiers for Land Cover Classification Using Landsat Multispectral Data," *IEEE Conference on Open Systems (ICOS2011)*, vol.3, pp.121-127, 2011, September 25 - 28, 2011, Malaysia.

Vapnik V., "Statistical Learning Theory," *Wiley Publishers Inc. New York*, pp.230-240, 1998.

Yanfeng Gu; Ying Liu; Ye Zhang, "A Soft Classification Algorithm based on Spectral-spatial Kernels in Hyperspectral Images," *International Conference on Innovative Computing, Information and Control, ICICIC '07*, pp.548, 2007.